\documentclass[11pt]{scrartcl}

\usepackage{fullpage}
\usepackage{authblk}
\usepackage{graphicx}
\usepackage{url}
\usepackage{caption}
\usepackage{subcaption}
\usepackage{algorithm,algpseudocode}
\usepackage{amssymb,amsmath}
\usepackage{multirow}
\usepackage{color}
\usepackage{hyperref}
\usepackage[square,numbers]{natbib}

\newcommand{\parbf}{\boldsymbol{\theta}}
\newcommand{\intset}{\mathbb{N}}
\newcommand{\best}[1]{\textbf{#1}}
\newcommand{\head}[1]{\centering #1}

\newcommand{\change}[1]{#1}
\newcommand{\rev}[1]{#1}

\begin{document}

\title{Unbiased Bayesian Inference for Population Markov Jump Processes via Random Truncations}



\author[1]{Anastasis Georgoulas}
\author[1]{Jane Hillston}
\author[1,2]{Guido Sanguinetti}

\affil[1]{School of Informatics, University of Edinburgh}
\affil[2]{SynthSys --- Synthetic and Systems Biology, University of Edinburgh}
\date{}



\maketitle

\begin{abstract}
We consider continuous time  Markovian processes where populations of individual agents interact stochastically according to kinetic rules. Despite the increasing prominence of such models in fields ranging from biology to smart cities, Bayesian inference for such systems remains challenging, as these are continuous time, discrete state systems with potentially infinite state-space. Here we propose a novel efficient algorithm for joint state / parameter posterior sampling in population Markov Jump processes. We introduce a class of pseudo-marginal sampling algorithms based on a random truncation method which enables a principled treatment of infinite state spaces. Extensive evaluation on a number of benchmark models shows that this approach achieves considerable savings compared to state of the art methods, retaining accuracy and fast convergence. We also present results on a synthetic biology data set showing the potential for practical usefulness of our work.
\end{abstract}

\section{Introduction}
Discrete state, continuous time stochastic processes such as Markov Jump Processes~(MJP) \cite{gardiner} are popular mathematical models used in a wide variety of scientific and technological domains, ranging from systems biology to computer networks. Of particular relevance in many applications are models where the state-space is organised according to a population structure (population Markov Jump Processes, pMJP): each state label corresponds to counts of individual entities in a number of populations (or species). These models are at the root of essentially all agent-based models, a class of models which is gaining increasing popularity in applications ranging from smart cities, to epidemiology, to systems biology. Despite their importance, solving inferential problems within the pMJP framework is challenging: the discrete nature of the system prevents the use of simple parametric distributions, and the size of the state space (which can be unbounded for open systems) effectively rules out analytical computations. At the same time, technological advances in areas as diverse as single cell biology and remote sensing are providing increasing amounts of data which can be naturally modelled as pMJPs, creating a pressing need for inferential methodologies.

In response to these developments, researchers in the statistics, machine learning and systems biology communities have been addressing inverse problems for MJPs using a variety of methods, from  variational techniques~\cite{opperVariational,cohn2010mean} to particle-based ~\cite{zechner2014scalable,hajiaghayi2014} and auxiliary variable sampling methods \cite{RaoTehLong}. Markov-chain Monte Carlo (MCMC) methods, in particular, offer a promising direction: while often computationally more intensive than variational methods, they provide asymptotically exact inference.
However, standard MCMC methods rely on likelihood computations, which are computationally or mathematically infeasible for pMJPs with a large or unbounded number of states. Such systems are commonplace in many applications, where one is often confronted with open systems where upper bounds on the numbers of agents are difficult to come by. As far as we are aware, current methods address this issue by arbitrarily truncating the state space according to pre-defined heuristics, offering no control over the error introduced by this procedure.

In this paper we present a novel Bayesian approach to posterior inference in pMJPs which solves these issues by adopting a pseudo-marginal approach based on random truncations, yielding both asymptotic exactness and computational improvements. We build on the auxiliary variable Gibbs sampler for finite state Markov Jump Processes (MJP) of~\cite{RaoTehLong}, significantly increasing its efficiency by leveraging the more compact representation of the kinetic parameters provided by the pMJP framework. We then present a novel formulation of the likelihood, which enables the deployment of a \emph{Russian Roulette}-like random truncation strategy as in~\cite{girolami2013playing,FilipponeICML15}. Based on this, we develop  a pseudo-marginal sampling approach for general pMJPs, obtaining two novel algorithms: a relatively straightforward Metropolis-Hastings pseudo-marginal scheme, and an auxiliary variable pseudo-marginal Gibbs sampler. We examine the performance of these algorithms in terms of  accuracy and efficiency on non-trivial case studies. We conclude the paper with a discussion of our contribution in the light of existing research and possible future directions in systems biology.

\section{Background }
\subsection{Population Markov Jump Processes}
\label{sec:stoch}
\begin{figure}
	\centering
	\includegraphics[scale=0.4]{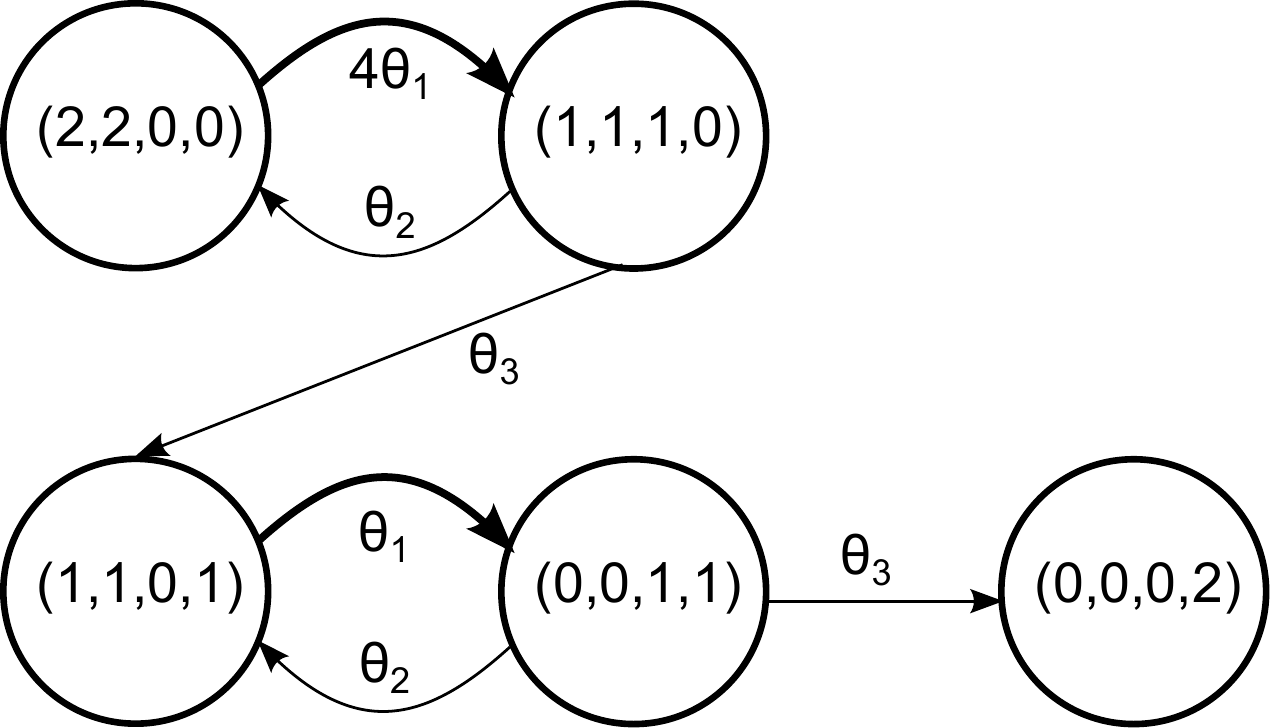}
	\caption{State-space of an example system. Arrows indicate transitions between states; the bolded transitions are ``instances" of the same reaction type, which updates the state by $(-1,-1,1,0)$ and occurs with rate $\theta_1 s_1 s_2$, where $s_1,s_2$ are the first and second components of the state.}
	\label{fig:exSpace}
\end{figure}

Population Markov Jump Processes are a particular type of Markov Jump Processes (also known as Population Continuous Time Markov Chains); they are continuous time stochastic processes whose discrete state
vector $\mathbf{s} = (n_1,n_2,\dots,n_M)$ gives the agent  counts of each of $M$ populations (or species) interacting through $R$ reaction channels. We will adopt here the language of chemical reactions to describe such processes, but the same considerations apply in general.
Reactions between individual agents (or molecules) happen as a result of random collisions, and each reaction changes the state by a finite amount, encoded in the {\it stoichiometry} of the system, corresponding to the creation/ destruction of a certain number of molecules.
Each reaction $i$ also has an associated \emph{kinetic law} giving its rate: this is generally of the form \begin{equation}
f_i(\mathbf{n}) = \theta_i \rho_i(\mathbf{n}),\label{kinLaw}\end{equation} 
where $\rho_i$ is a {\it fixed} function of the state $\mathbf{n}$, while $\theta_i$ are (usually unknown) kinetic parameters. Therefore, while in a general MJP there can be a parameter associated with each possible transition, in pMJPs the dynamics are captured more succinctly by a single parameter per reaction.
A schematic of a simple pMJP is given in Figure~\ref{fig:exSpace}, where it can be seen that the same reaction can correspond to multiple transitions in the state-space of the process, all of which follow the same kinetic law and incur the same update to the state.

The time evolution of the process marginals is given by the Chemical Master Equation (CME):
\begin{equation}
\label{eq:CME}
\frac{dp_i(t)}{dt} = \sum_{j \neq i} p_j(t) a_{ji} - p_i(t)\sum_{j \neq i} a_{ij}
\end{equation}
where $p_i(t)$ is the probability of being in state $i$ at time $t$ and $a_{ij}$ is the rate of jumping from state $i$ to state $j$, which for pMJPs is known from the kinetic law.


For finite state-spaces, one can gather the transition rates $a_{ij}$ in the \emph{generator matrix} $\mathbf{A}$, and the CME can be solved analytically as:
\begin{equation}
\label{eq:exp}
\mathbf{p}(t) = \mathbf{p}(0) e^{\mathbf{A}t}
\end{equation}
This solution can be computationally intensive, even with the use of specialized algorithms like~\cite{exponentialAction}.

\subsection{Uniformisation and inference}
An alternative approach to solve the CME is given by {\it uniformisation}~\cite{jensen1953markoff}, a  well-known technique for the transient analysis of Markovian systems, used widely in fields like performance modelling.
Given a MJP with generator $\mathbf{A}$, uniformisation constructs a discrete time Markov chain by imposing a common exit rate $\gamma$ for all states. For this procedure to be consistent, $\gamma$ must be no less than the highest exit rate among all states. The resulting {\it uniformised} system is then faster than the original, in the sense that transitions occur at a higher rate. To compensate for this and maintain the behaviour of the original MJP, virtual jumps must be added from each state to itself. 
This results in a discrete time system with transition probability matrix $B = \frac{1}{\gamma} \mathbf{A} + I$, in which likelihood computations are standard. 
In this discrete time system, the waiting time before a jump occurs now follows an exponential distribution with rate $\gamma$, regardless of the current state. The probability of jumping from state $i$ to another state $j$ is $\frac{a_{ij}}{\gamma}$, but it is now also possible to remain in $i$ after the jump, with probability $1 - \frac{1}{\gamma}\sum_{j \neq i} a_{ij}$.
Jensen's classical result \cite{jensen1953markoff} then guarantees that all the time-marginals of the discrete time process match those of the continuous time chain.

Uniformisation has previously been exploited by~\cite{RaoTehLong} to draw posterior samples from a MJP conditioned on a set of observations. The idea is to construct a discrete time chain using uniformisation, sample a trajectory (including self-loops) and run a standard forward filtering-backward sampling (FFBS) algorithm on it. This gives a new trajectory which, when self-jumps are removed, is a sample from the posterior process. This path-sampling algorithm can be alternated with Gibbs updates to jointly sample transition probabilities; in \cite{RaoTehLong} this is accomplished by choosing conjugate Dirichlet priors on each entry of the generator matrix, resulting in potentially many parameters with consequent storage/ computational issues.


\section{Unbiased sampling for pMJPs}
\subsection{Efficient Gibbs sampling for finite state pMJPs}
\label{sec:inf}
The special structure of pMJP systems implies considerable inferential savings over the generic Gibbs sampler \cite{RaoTehLong}. In particular, the functional form of the kinetic law associated with the $i$th reaction, $f_i(\mathbf{s}) = \theta_i \rho_i(\mathbf{s})$, suggests a different conjugate prior for the parameters $\theta_i$, which greatly simplifies the parameter sampling steps within the Gibbs sampler.


Let $(S,T)$ be a full trajectory sampled from the uniformised conditional posterior in a Gibbs step, where $S = (s_0,s_1,\dots,s_K)$ is the sequence of states at times $T = (t_0,t_1,\dots,t_K)$. Let $u_k$ denote the reaction at time $t_{k+1}$, as inferred\footnote{We assume each reaction has a distinct update vector.} from inspection of $s_k$ and $s_{k+1}$. From Section~\ref{sec:stoch}, we know that the total rate of exiting state $s_k$ is $r_k = \sum_{i = 1}^R \theta_i \rho_i(s_k)$. Since the waiting time between jumps is exponentially distributed in a MJP, this gives 
\[
p(t_{k+1} \mid t_k, s_k) = r_k e^{-\Delta t_k r_k} \text{ , where } \Delta t_k = t_{k+1} - t_k
\]
The probability of the next state being $s_{k+1}$ is $\tfrac{\theta_{u_k} \rho_{u_k}(s_k)}{r_k}$. The total likelihood is then
\begin{align*}
L(\parbf) &= p(S,T \mid \parbf) =p(S \mid \parbf) p(T \mid S, \parbf) \\
&= \prod_{k = 0}^{K-1} \frac{\theta_{u_k} \rho_{u_k}(s_k)}{r_k} 
r_k e^{-\Delta t_k r_k} \\
&= \prod_{k = 0}^{K-1} \theta_{u_k} \rho_{u_k}(s_k) e^{-\Delta t_k r_k}
\end{align*}
Let each parameter be Gamma-distributed \textit{a priori}:
\[p(\theta_i) = \frac{b_i ^{a_i}}{\Gamma(a_i)}\theta_i ^{a_i - 1} e^{-b_i \theta_i}\]
We then have:
\begin{align}
p(\theta_i \mid S,T) &\propto p(\theta_i) p(S,T \mid \parbf) \nonumber \\
&\propto \theta_i ^ {a_i + N_i - 1} e^{-b_i -\sum_{k = 0}^{K-1} \Delta t_k \rho_i(s_k) \theta_i}
\label{eq:postPar}
\end{align}
Therefore, conditioned on the trace, the parameters are again Gamma-distributed with shape $a_i + N_i$ and rate $b_i + \sum_{k = 0}^{K-1} \Delta t_k \rho_i(s_k)$, where $N_i$ is the number of times the $i$th reaction type is observed in the trace. Hence, we have \emph{exact} Gibbs updates for the kinetic parameters; notice that, since we have a single parameter for each reaction, the number of parameters to be sampled is often orders of magnitude lower than the number of parameters sampled in \cite{RaoTehLong} (one per possible state transition), yielding computational and storage savings.

\subsection{Unbounded state-spaces}
Many pMJPs of practical interest describe open systems with infinite state-spaces, which are not amenable to uniformisation.  A plausible solution would be to truncate the system, possibly using methods such as in~\cite{fsp} to quantify the error. However, any such bound would be dependent on the unknown parameters, and in order to achieve acceptable performance we may need to still retain very large state spaces. An alternative approach may be to introduce random truncations in such a way as to obtain an unbiased estimator of the likelihood, which can be used in a pseudo-marginal MCMC scheme~\cite{andrieu2009pseudo,Beaumont1139}. We describe here two algorithms based on random truncations, a simple Metropolis-Hastings \mbox{(M-H)} sampler directly targeting the marginal likelihood, and a Metropolized auxiliary variable Gibbs sampler. 

\subsubsection{Expanding the likelihood}
We start by describing a formulation of the likelihood in the pMJP setting as an infinite series.
The basic idea is to decompose the space of process trajectories into a nested sum over subspaces of trajectories which differ by at most $N$ from the observations. We can then define a generator matrix on each of these finite state-space systems and compute transient probabilities using~(\ref{eq:exp}). We now explicitly define the terms in this expansion of the likelihood. For simplicity, we focus on deriving the likelihood for a single, noiseless observation $(t',s')$ in a one-dimensional process, assuming the state at time $0$ is known to be \change{$s \in \intset$}. Due to the Markovian nature of the process, the actual likelihood will be given by a product of such terms.  If we write $s^u = \max(s,s')$, we have:
\begin{align}
p(s' \mid s,\parbf) &= \sum_{N = 0}^{\infty} p(s', \max{(s_{0:t'} - s^u)} = N \mid x, \parbf) \nonumber \\
&\equiv \sum_{N=0}^{\infty} p^{(N)}(s',s)
\label{eq:like}
\end{align}

\change{The notation $s_{0:t}$ indicates all values of the process in the time interval $[0,t]$ and is used here as follows: $\max{(s_{0:t})} = N$ means that the maximum value of the process in the interval $[0,t]$ is $N$. Similarly, $\max{(s_{0:t})} \le N$ means that the process does not exceed the value $N$ during $[0,t]$.}

Note that the constraint on the maximum of $s_{0:t'} - s^u$ does not simply define a state-space, but constrains us to consider only those trajectories that actually achieve a ``dispersal'' of $N$. If we define
\[f^{(N)}(s',s) = p(s', \max{(s_{0:t} - s^u)} \le N \mid x,\parbf )\]
then each term of the series can be decomposed as:
\begin{equation}
p^{(N)}(s',s) = f^{(N)}(s',s) - f^{(N-1)}(s',s)
\label{eq:decomp}
\end{equation}

These sub-terms are now the transient probability for a finite state-space pMJP, and can be computed using Equation~\ref{eq:exp}. Any number of them are computable but, naturally, the whole sum cannot be computed in finite time. It can, however, be estimated in an unbiased way.

\subsubsection{Random truncations}
\label{sec:rr}
Assume we wish to estimate an infinite sum
\[f = \sum_{N = 0}^\infty f_N\]
where each term $f_N$ is computable. One way of approximating the sum is to pick a single term $f_k$, where $k$ is chosen from any discrete distribution with mass $p_0,p_1,\dots$. We can immediately see that $\hat{f} = \frac{f_k}{p_k}$ has expectation $E[\hat{f}] = \sum_{N=0}^\infty \frac{f_N}{p_N} p_N = f$ and is therefore an unbiased estimator of the infinite sum. An issue with this approach is that, depending on the choice of distribution $p_i$, the variance of $\hat{f}$ might be very large, even infinite.


A reduced variance estimator can be obtained by approximating $f$ with a partial sum up to order $N$, weighted appropriately. The number of terms is chosen randomly: at every term $j$, a random choice is made: there is a probability $q_j$ of stopping the sum, otherwise we continue to form iteratively the partial sum $\hat{f} = \sum_{N=0}^{j} \frac{f_N}{p_N}$, where $p_N = \prod_{j=1}^{N-1}(1-q_j)$. This scheme, imaginatively termed  \emph{Russian Roulette sampling} \cite{girolami2013playing}, can also be shown to yield an unbiased estimator of $f$. 

\subsubsection{Metropolis-Hastings sampling}
Applying this random truncation strategy to the expansion in~\eqref{eq:like} produces an unbiased estimator.
Such estimates can be obtained for every interval between successive observations; since they are independent, their product will be an unbiased estimate of the likelihood under all the observations. \change{Note that each summand in \eqref{eq:like} is a probability, and is therefore non-negative. Thus, we avoid the problems of possibly negative estimators; this positivity is important, as non-positive estimators may result in a large or infinite variance. It is worth remarking that the term for $N=0$ corresponds to a space that includes the observations at both ends of the time interval, and hence will already include a significant contribution of probability mass towards the likelihood.}

\change{The same approach is easily extended to higher dimensions, where the states are vector of integers, by adapting the notation: $\max{(\mathbf{s}_{0:t})} \le N$ means that the value in \emph{any} dimension does not exceed $N$ in the given interval, whereas $\max{(\mathbf{s}_{0:t})} = N$ now means that a value of $N$ is not exceeded in any dimension during $[0,t]$, and that it is achieved in at least one dimension.}

This procedure directly gives rise to a pseudo-marginal M-H algorithm, where the likelihood term is approximated by the unbiased estimate obtained as described above.
We refer to this as \emph{Algorithm 1} and examine its performance in the next section.

For our purposes, we choose a $q_n$ sequence such that the probability of accepting a term decreases geometrically; specifically, we use $q_n = 1 - a(1-q_{n-1})$, with $q_0 = 0$ and $a = 0.95$. We note that, since all terms in the series are non-negative and tend to $0$, we can make use of a result from~\cite{girolami2013playing} to show that the variance of the estimator is finite. We show an empirical analysis of the variance in Section~\ref{sec:var} that validates our choice of $q_n$ and indicates that performance is robust with respect to the choice of the particular stopping distribution.

\subsubsection{Modified Gibbs sampling}
An alternative approach is to incorporate the truncation in the Gibbs sampler described in Section~\ref{sec:inf}. The difficulty is that there is no direct way to sample trajectories without a bound on the state-space, as the uniformisation sampler requires a finite number of states. To work around this limitation, we propose to sample a truncation point, then draw a trajectory and parameters for this state-space as in Section~\ref{sec:inf}. Since we are no longer sampling from the true conditional posterior over trajectories, but rather are also conditioning on the chosen truncation, we are no longer able to accept every trajectory and parameter sample drawn. Instead, we must introduce an acceptance ratio that will ensure we are sampling from an unbiased estimate of the true conditional posterior. We refer to this as \emph{Algorithm 2}; the following is a summary of the procedure to form a new sample $(\parbf_{t+1},S_{t+1})$ from the current state $(\parbf_t,S_t)$ of the chain, given a set of observations $O$: 

\begin{enumerate}
	\item\label{stepPar} Sample $\parbf^* \mid S_t$, as detailed above.
	\item Sample $S^* \mid O,\parbf^*$:
	\begin{enumerate}
		\item\label{stepTrunc} Choose a truncation point $m^*$, defining a finite state-space.
		\item\label{stepTrace} Run the FFBS algorithm to draw $S^*$.
	\end{enumerate}
	\item Calculate the acceptance ratio $\alpha$:
	\begin{enumerate}
		\item\label{stepProb1} Compute $p^{(t+1)}(S^* \mid \parbf^*,O)$ and $p^{(t)}(S^* \mid \parbf^*,O)$, the conditional posterior probabilities of the \emph{new} trajectory under the \emph{new} and \emph{old} truncations.
		\item\label{stepProb2} Compute $p^{(t+1)}(S_{t} \mid \parbf^*,O)$ and $p^{(t)}(S_{t} \mid \parbf^*,O)$, the conditional posterior probabilities of the \emph{old} trajectory under the \emph{new} and \emph{old} truncations.
		\item Set $\alpha = \frac{p^{(t+1)}(S^* \mid \parbf^*,O) p^{(t+1)}(S_{t} \mid \parbf^*,O)}{p^{(t)}(S_{t} \mid \parbf^*,O) p^{(t)}(S^* \mid \parbf^*,O)}$.
	\end{enumerate}
	\item\label{stepAccept} With probability $\min(\alpha,1)$, accept the new sample and set $(\parbf_{t+1},S_{t+1}) = (\parbf^*,S^*)$; otherwise, set $(\parbf_{t+1},S_{t+1}) = (\parbf_t,S_t)$
\end{enumerate}

\change{Note that the analysis from Section~\ref{sec:inf} giving the conditional posterior of the parameters (Equation~\eqref{eq:postPar}) still holds and is not affected by the truncation. Step~\ref{stepPar} is therefore performed following~\eqref{eq:postPar}.
	In Step~\ref{stepTrunc}, we follow the Rusian Roulette methodology as in Section~\ref{sec:rr} and take $m^*$ to be the number of terms before the truncation stops. In the scheme used in our experiments, the probability of taking an additional term follows a geometric distribution, as with the previous algorithm. Based on this truncation point $m^*$, we can define a state-space \[\mathcal{S} = \left\{ (x_1,x_2,\dotsc,x_M) \in \intset^M \mid x_i \le y^*_i + m^* \right\}\] where $y^* = (y_1,\dotsc,y_M)$ is a vector of the maximum values observed in each dimension.
	The method of Section~\ref{sec:inf} can then be used to sample a trajectory  (Step~\ref{stepTrace}) in this finite state-space.}

Steps~\ref{stepProb1} and~\ref{stepProb2} involve the computation of probabilities which can be performed via the forward-backward algorithm on the appropriate state-spaces. So far in this paper, the algorithm has been used to sample a new path from the process, but it can easily be adapted to calculate the probability of a given path, as shown in the algorithm outline below.

\change{In the following, we assume we have $N$ observations $y_i$ at time points $t_i$, $i = 1,\dotsc,N$. For a finite state-space $\mathcal{S}$, we denote with $\mathcal{S}_k$ the $k$-th state of the space, according to some arbitrary order. The forward and backward messages are vectors of size $\left| \mathcal{S} \right|$, and there is one such message for each observed time point. $\mathbf{a}^{(i)}$ denotes the forward message at the $i$-th time point $t_i$; its $k$-th element is
	\[
	a_k^{(i)} = p(y_1,\dotsc,y_{i-1},\mathcal{S}_k)
	\]
	that is, the joint probability of the observations prior to $t_i$ and the state at $t_i$ being $\mathcal{S}_k$. Similarly, the backward messages $\mathbf{b}^{(i)}$ has elements:
	\[
	b_k^{(i)} \propto p(\mathcal{S}_k \mid y_{i},\dotsc,y_{N})
	\]
	and so the probability of the observed time-series can be computed from the $\mathbf{b}^{(i)}$. This is a slightly different than the usual formulation of the forward-backward algorithm, and necessitates the computation of the forward messages $\mathbf{a}^{(i)}$ first. The messages can be computed recursively as shown in~\cite{RaoTehLong}.}

\begin{algorithm}
	\caption*{\change{Forward-backward algorithm}}
	\label{alg:fb}
	\begin{algorithmic}[1]
		\Require Observations $Y = (y_1,\dotsc,y_N)$, finite state-space $\mathcal{S}$, parameters $\parbf$
		\Ensure Probability of $Y$
		\State Compute transition probabilities $p_{kl}$ between states in $\mathcal{S}$ based on $\parbf$
		\For{$i = 1..N$} 
		\State Compute forward message $\mathbf{a}^{(i)}$
		\EndFor
		\State Initialise $p \gets 1$
		\For{$i = N..1$}
		\State Compute backward message $\mathbf{b}^{(i)}$
		\State Find index $k$ of observation $y_i$ in $\mathcal{S}$
		\State $p = p \cdot b_{k}^{(i)}$
		\EndFor
		\State \Return $p$
	\end{algorithmic}
\end{algorithm}

These probabilities computed this way are then used in the acceptance ratio $\alpha$ (Step~\ref{stepAccept}).
As noted above, the acceptance step is necessary because we are not proposing trajectories from the exact conditional posterior. Instead, the truncation we impose gives an estimate of the correct proposal distribution $p(S^* \mid \parbf^*,O)$, and the ratio compensates for this estimate. Note that, if we could draw trajectories from the whole state-space without truncating it, the terms in $\alpha$ would cancel out, giving standard Gibbs sampling with acceptance rate of 1.

It is important to observe that this auxiliary variable Gibbs sampler actually targets the joint posterior distribution of parameters and trajectories. As such, it provides richer information than the M-H sampler (which directly targets the parameter posterior), but may be less effective if one is solely interested in parameter inference. The performance can also be affected by computational factors, particularly the costs of drawing sample trajectories (which was not needed \change{in Algorithm~1, where we compute the likelihood by matrix exponentiation}). In general, such costs will be model- and data-dependent, so that some initial exploration may be advisable before deciding which algorithm to use.

\section{Results}
\label{sec:results}
This section describes the experimental validation of our approach. The experiments were performed on MATLAB implementations of the algorithms described in the previous section\footnote{available at \url{https://github.com/ageorgou/roulette}}. \change{The M-H proposals for Algorithm~1 were Gaussian, with variances tuned using trial runs. In the same algorithm, matrix exponentiation was performed using the method of~\cite{exponentialAction}, with the code that the authors have made available. Unless otherwise noted, the Russian Roulette truncation used in the experiments was chosen so as to yield 5.6 terms on average.}

\subsection{Variance of the estimator}
\label{sec:var}
Before showing how our algorithms perform against the state of the art, we present empirical evidence that our Russian Roulette-style truncation approach produces estimators with low variance, an issue that has recently received attention in pseudo-marginal methods~\cite{Doucet07032015,sherlock2015}. \change{In order to achieve estimators of low variance, the tails of the distribution of the number of terms taken must match those of the sequence being approximated~\cite{rhee2013unbiased} (or the estimator is likely to ignore significant terms). To our knowledge, there are no established results on the behaviour of transient probabilities in general pMJPs as the state-space grows. Our approach is to use a geometric truncation distribution, which is well known (\cite{kleinrock}) to arise as a steady-state distribution of simple pMJPs such as queueing systems, and might thus be a plausible candidate distribution. Our focus in this section is to provide an empirical evaluation of our method.}
Additionally, we show that the estimator is robust to the choice of the particular stopping distribution $q_n$ used in the truncation scheme. To verify this, we considered three different $q_n$ sequences, applied to the predator-prey model described in Section~\ref{sec:bench}. For clarity, we write $\bar{q}_n \equiv 1 - {q}_n$, the probability of continuing at term $n$. All schemes were of the form $\bar{q}_n = a \bar{q}_{n-1}$ with $\bar{q}_0 = 1$ and $a \in \{0.95,0.75,0.2 \} $, respectively yielding 5.6, 2.4 and 1.2 terms on average.
For each scheme, we calculated 1000 estimates of the transition probabilities between observations, obtaining estimates of the log-likelihood and computing its mean and variance. This was repeated for 10 different parameterizations of the model. It can be seen (Table~\ref{tab:var}) that the variance of the estimator (measured as the coefficient of variation of the log-likelihood, due to small values) is consistently low. This validates our approach and indicates that the stopping distribution does not critically affect performance and therefore does not require fine-tuning.
\begin{table}
	\begin{center}
		\begin{tabular}{r|r|r}
			\multicolumn{1}{c|}{\head{$a = 0.95$}} & \multicolumn{1}{c|}{\head{$a = 0.75$}} & \multicolumn{1}{c}{\head{$a = 0.2$}}\\
			\multicolumn{1}{c|}{\head{(5.6 terms)}} & \multicolumn{1}{c|}{\head{(2.4 terms)}} & \multicolumn{1}{c}{\head{(1.2 terms)}}\\
			\hline
			0.0002 & 0.0008 & 0.0223\\
			0.0051 & 0.0151 & 0.0344\\
			0.0003 & 0.0013 & 0.0245\\
			$< 10^{-4}$ & $< 10^{-4}$ & 0.0016\\
			$< 10^{-4}$ & $< 10^{-4}$ & 0.0008\\
			0.0005 & 0.0021 & 0.0109\\
			$< 10^{-4}$ & $< 10^{-4}$ & $< 10^{-4}$\\
			0.0003 & 0.0014 & 0.0223\\
			0.0002 & 0.0011 & 0.0073\\
			0.0002 & 0.0009 & 0.0077\\
		\end{tabular}
	\end{center}
	
	\caption{Coefficient of variation for the log-likelihood, estimated from 1000 samples for the LV model, under three truncation schemes (varying $\alpha$) and ten parameter configurations (Section~\ref{sec:var})}\label{tab:var}
\end{table}

\begin{figure}
	\begin{subfigure}[b]{.4\linewidth}
		\includegraphics[scale=0.38]{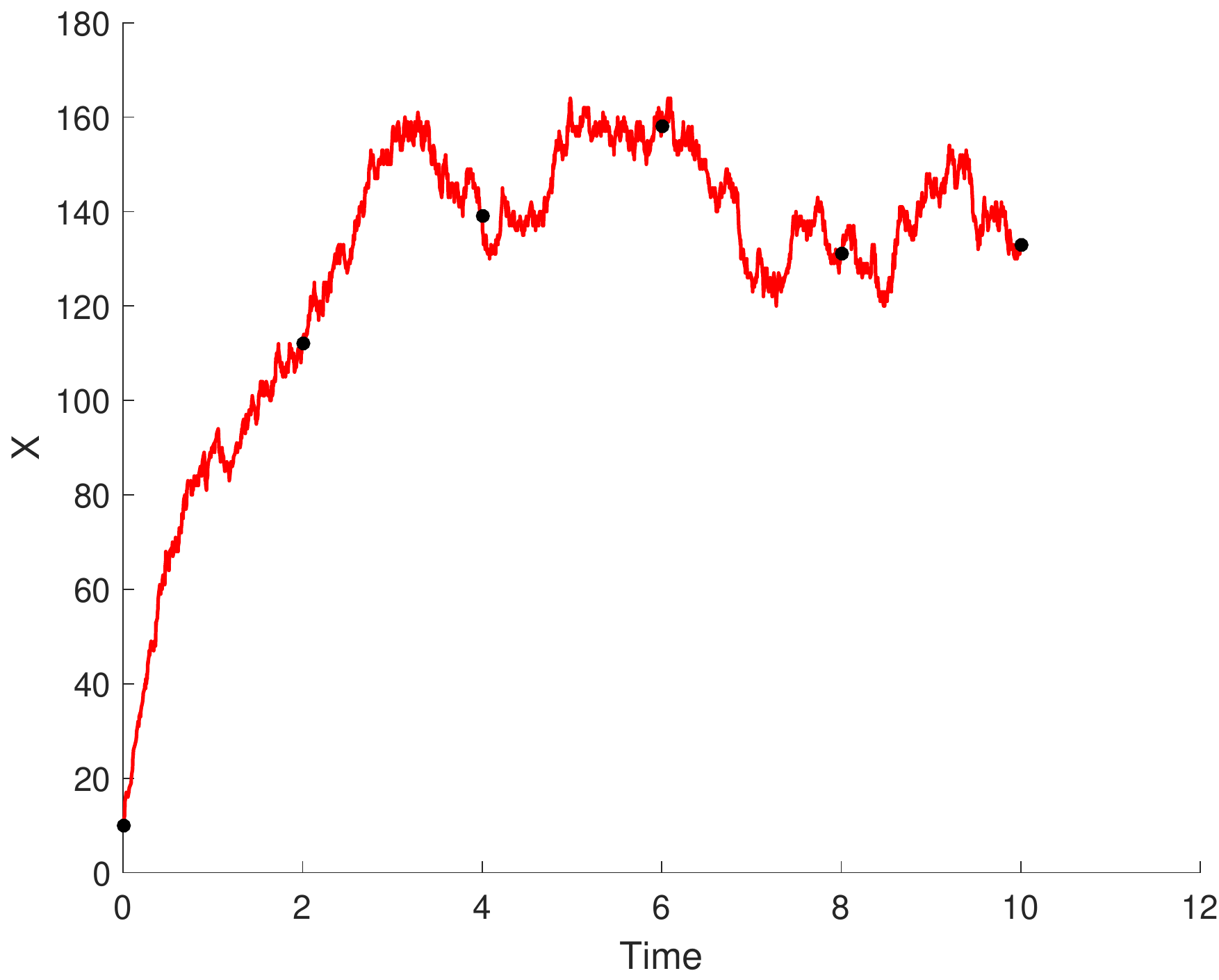}
		\caption{}
		\label{fig:sticking_obs}
	\end{subfigure}
	\hfill
	\begin{subfigure}[b]{.4\linewidth}
		\hspace{-2.7em}
		\includegraphics[scale=0.4]{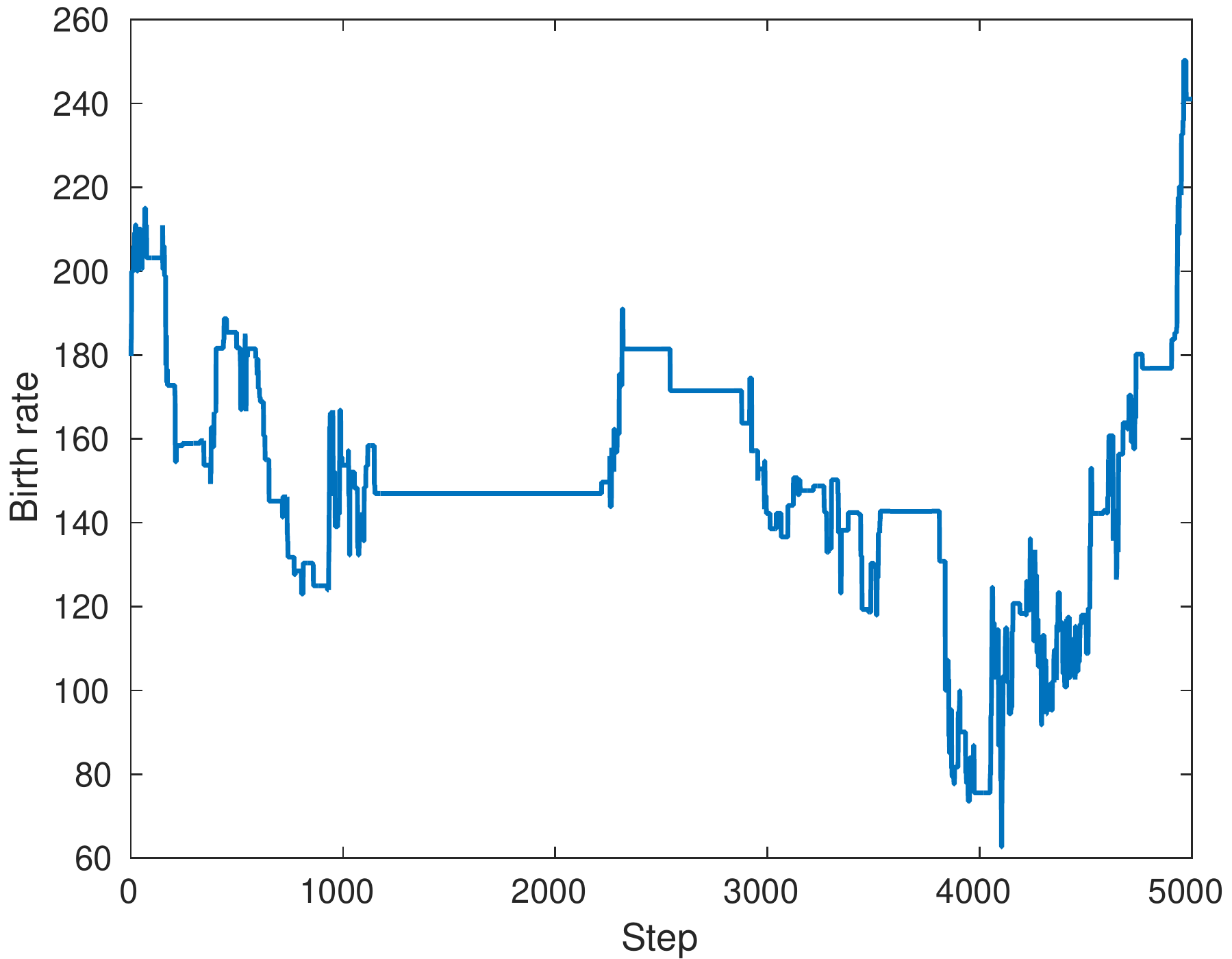} 
		\caption{}
		\label{fig:sticking}
	\end{subfigure}
	\caption{\change{(a) Full trace (continuous line) and observations (dots) used in the birth-death process example; (b) Parameter samples for the birth rate using Algorithm 1, illustrating undesirable ``sticking'' behaviour when taking 5.6 terms on average}}
\end{figure}

\change{An intuitive explanation for this comes from remembering that the ``base'' space (corresponding to the first term in the expansion) comprises all states between consecutive observations. Often, this is large enough that there is a substantial probability of the process remaining within or around it. Hence, even with a few terms, we are capturing a large part of the probability mass, and obtaining good estimates.
	We expect our estimator to have low variance if the process does not change radically between the observation times. It is possible, however, to find situations where the truncation strategy needs many terms in order to yield good performance. This is more likely to occur if the process is very sparsely observed, or if it is highly volatile. In both cases, the observations may not be very indicative about the behaviour of the process during the interval under consideration, therefore only taking few terms may produce inaccurate estimates of the true likelihood. This situation is also likelier when high counts are involved, in which case other proposed solutions are more appropriate (discussed in Section~\ref{sec:related}).
	
	To illustrate this, we considered the example of a birth-death process involving a single species, $X$, with a constant birth rate of $150$ and a death rate of $X$. From an initial value of $X=10$, we simulated the system and used the values at 5 time points (Figure~\ref{fig:sticking_obs}). The three truncation schemes described above did not yield accurate estimates, even when taking 5 terms on average. With a stopping scheme $\bar{q}_n = 0.99 \bar{q}_{n-1}$ (corresponding to 12.5 terms on average), we were able to get good estimates of the true probabilities. The more aggressive truncation schemes display higher variance and could cause problems when their estimates are used in Algorithm 1: when taking 5.6 terms on average, the variance causes the sampling chain to ``stick'', as seen in Figure~\ref{fig:sticking}.}

\rev{As a way of improving the behaviour of the sampler, we examined the use of the so-called Monte Carlo within Metropolis (MCWM) pseudomarginal variant~\cite{Beaumont1139}, in which the estimate of the likelihood of the current state of the chain is recomputed at every step. This can potentially alleviate the ``sticking'' problem and lead to better mixing, but at the cost of making the resulting chain sample from an approximation instead of the true posterior. Experiments on the predator-prey model of Section~\ref{sec:bench} showed that there was no noticeable improvement in either the number of steps needed to reach convergence or the acceptance rate when using MCWM. This, in addition to the bias introduced and the additional computational burden from re-estimating the likelihood, leads us to believe that in this case there is no benefit from using MCWM.}

\begin{figure}
	\centering
	\includegraphics[scale=0.4]{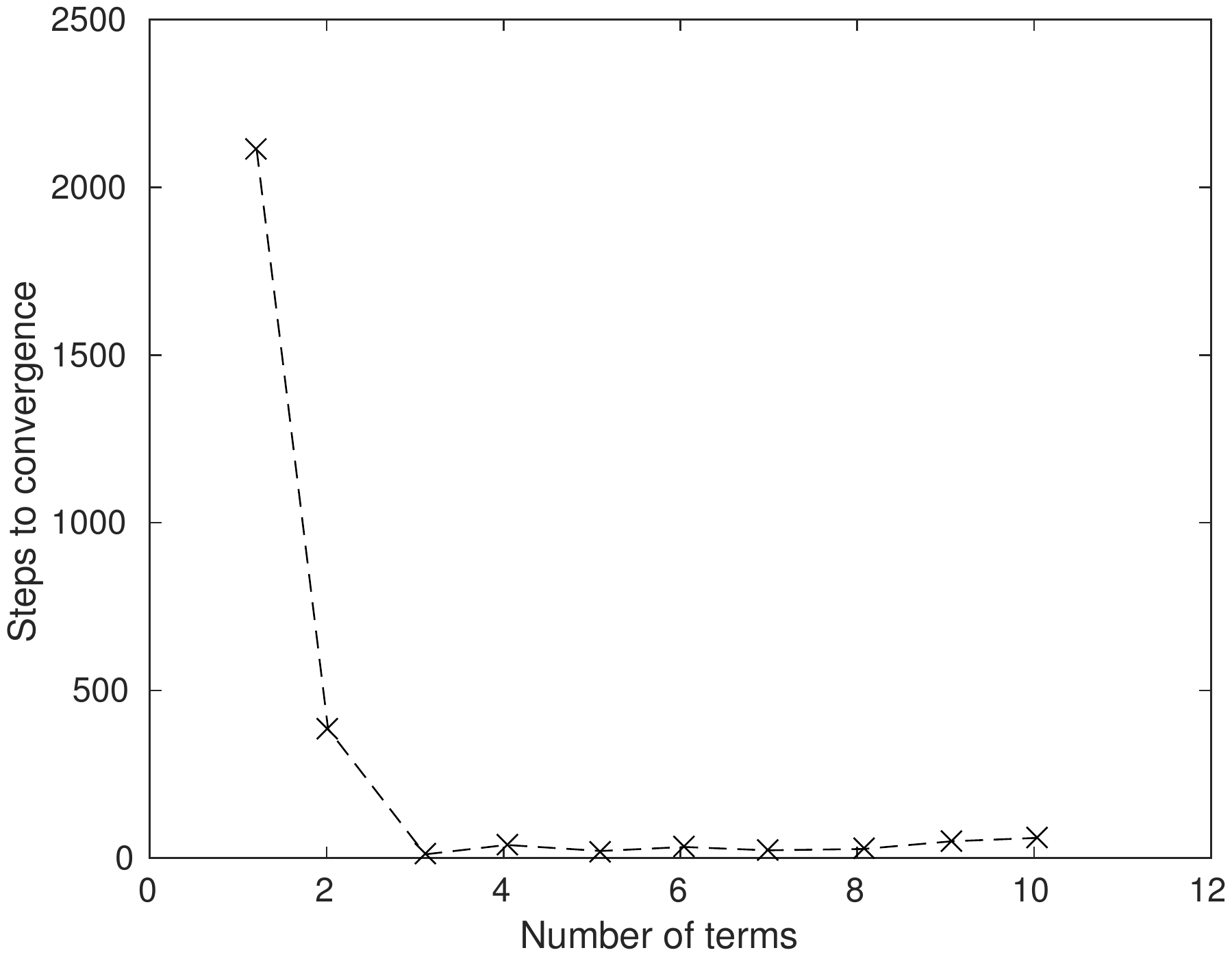}
	\caption{Steps until convergence for different stopping distributions (results shown for one parameter of the LV model).}
	\label{fig:variableConv}
\end{figure}

\rev{To further study of the impact of the choice of truncation distribution, we examined how it affects convergence. We tried ten different stopping distributions $q_n$ of the form described above, chosen so that they produce $1,2,\dotsc,10$ terms on average. For each of them, we measured the steps required for convergence, as described in the next section. Overall, we found that taking more terms generally leads to faster convergence (Figure~\ref{fig:variableConv}). This indicates that the variance of the estimates decreases when taking more terms.}

\subsection{Benchmark data sets} \label{sec:bench}
We now assess the performance the two algorithms described in the previous section as well as the Gibbs sampler based on uniformisation (Section~\ref{sec:inf}). We could not run the original Gibbs sampler of \cite{RaoTehLong} as the high number of parameters (one per state) swiftly led to storage problems.
We first compared the performance of the three methods on two widely used pMJP models:
\paragraph{Lotka-Volterra (LV) model}
This predator-prey system involves four types of reactions, representing the birth and death of each species, and is a classic model in ecology and biochemistry. Truncated LV processes have been studied in previous work~(\cite{opperVariational},\cite{boys2008bayesian}), making it an attractive candidate for evaluating our approach.
\[
\begin{array}{lcll}
X + Y & \rightarrow& 2X + Y & \text{at rate } \theta_1 X Y\\
X &\rightarrow& \varnothing & \text{at rate } \theta_2 X\\
Y &\rightarrow& 2Y & \text{at rate } \theta_3 Y\\
X + Y &\rightarrow& X & \text{at rate } \theta_4 X Y 
\end{array} 
\]
We start from an initial state of 7 predators and 20 prey. When a finite state-space is required, we impose a maximum count of 100 for each species, as in previous work.

\paragraph{SIR epidemic model}
A commonly-used model of disease spreading (see e.g. \cite{anderson1991infectious}), where the state comprises three kinds of individuals: S(usceptible), I(nfected) and R(ecovered). We examine two variants of the model, a finite version where the total population is constant:
\[
\begin{array}{lcll}
S + I &\rightarrow& 2I & \text{at rate } \theta_1 S I\\
I &\rightarrow& R & \text{at rate } \theta_2 I\\
\end{array}
\]
and an infinite state variant where new individuals can join the S population with unknown arrival rate:
\[
\begin{array}{lcll}
\varnothing  &\rightarrow& S & \text{at rate } \theta_3\\
\end{array}
\]
The initial state in both cases is $(S,I,R) = (10,5,0)$. For the finite-state version, this gives a state-space of 121 states. For the infinite case, we chose a truncation with upper limit $(28, 33, 33)$, corresponding to 18 new arrivals in the system. To see this, note that the number of arrivals in a time interval of duration $T$ is Poisson-distributed, with mean $\theta_3T$. We used the final observation time and the prior mean of $\theta_3$, and chose the 95-percentile of the distribution governing the new arrivals. In broad terms, this means our truncation will accommodate new arrivals with 95\% probability.

Table~\ref{tab:results} summarises our evaluation results across the models considered; the metrics we use are total computational time for 5000 samples, mean relative error in parameter estimates (using the posterior mean as a point estimate), Effective Sample Size (ESS) per minute of computation, and number of iterations to convergence, defined as Potential Scale Reduction Factor (PSRF) $<1.1$ \cite{gelman2014bayesian}.

\begin{table}
	\begin{center}
		\begin{tabular}{|c|c|c|c|c|}
			\hline
			& & Gibbs & Alg.1 & Alg. 2\\
			\hline
			\multirow{4}{*}{LV}& Time & 1011min & 55min & \best{29min} \\
			\cline{2-5}
			& Error & 14\% & \best{10.66}\% & 12.75\%\\
			\cline{2-5}
			& ESS/min & 0.63 & 0.67 & \best{4.5}\\
			\cline{2-5}
			& Iter. & \best{24} & 1314 & 180\\
			\hline			
			\hline
			\multirow{4}{*}{SIR finite} & Time  & \best{1min} & 10min & 4min \\
			\cline{2-5}
			& Error & 2.24\% & 13.17\% & \best{2.13\%} \\
			\cline{2-5}
			& ESS/min & \best{1752.13} & 63.01 & 422.72\\
			\cline{2-5}
			& Iter. & \best{13} & 33 & 27\\
			\hline
			\hline
			\multirow{4}{*}{SIR infinite}& Time & 1585min & \best{291min} & 666min \\
			\cline{2-5}
			& Error & 31.6\% & 25\% & \best{24.3\%} \\
			\cline{2-5}
			& ESS/min & 0.45 & \best{2.6} & 0.23 \\
			\cline{2-5}
			& Iter. & \best{5} & 65 & 136 \\
			\hline
		\end{tabular}
		
	\end{center}
	\caption{Performance of the various algorithms tested. Metrics are averaged over all parameters. Experiments were performed on a 24-core Xeon E5-2680 2.5GHz, to accommodate the increased memory requirements of some cases.}
	\label{tab:results}
\end{table}

%

\begin{figure*}
	\begin{subfigure}[b]{.45\linewidth}
		\centering
		\includegraphics[scale = 0.5]{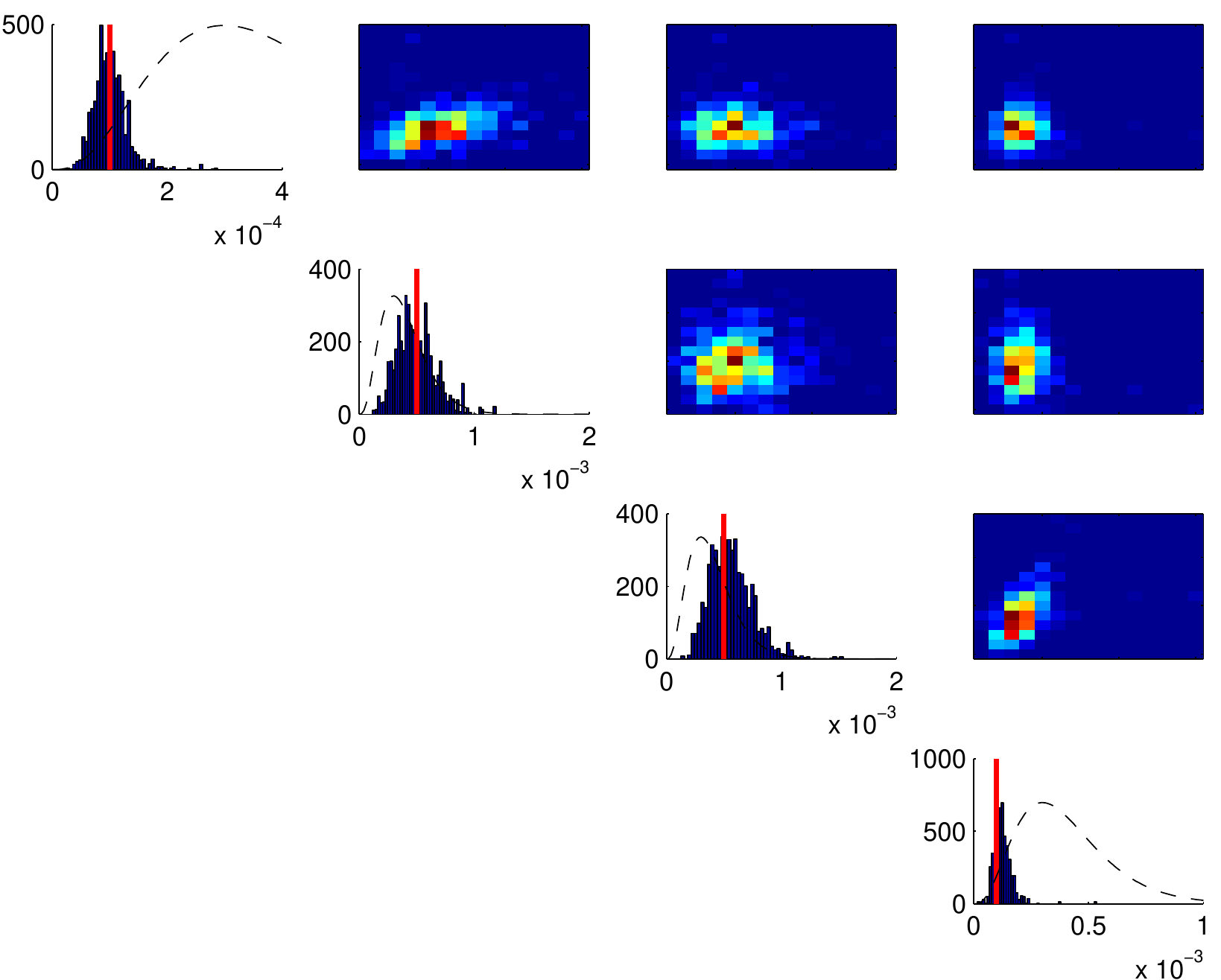}
		\caption{}
		\label{fig:panels}
	\end{subfigure}
	\hfill
	\begin{subfigure}[b]{.45\linewidth}
		\hspace{3em}
		\begin{subfigure}{.45\linewidth}
			\centering
			\includegraphics[scale=0.3]{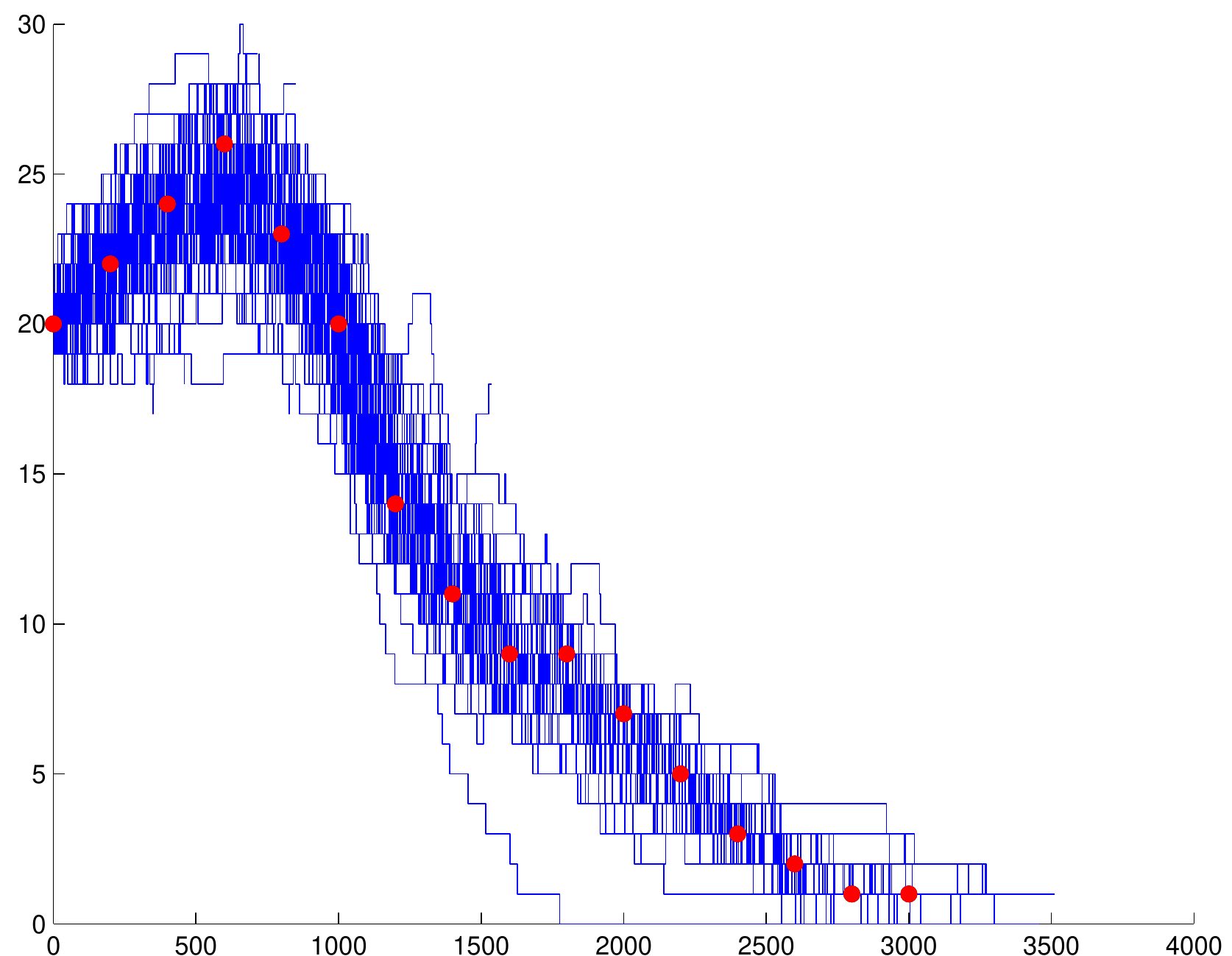}
		\end{subfigure}
		
		\medskip
		\hspace{3em}
		\begin{subfigure}{.45\linewidth}
			\centering
			\includegraphics[scale=0.3]{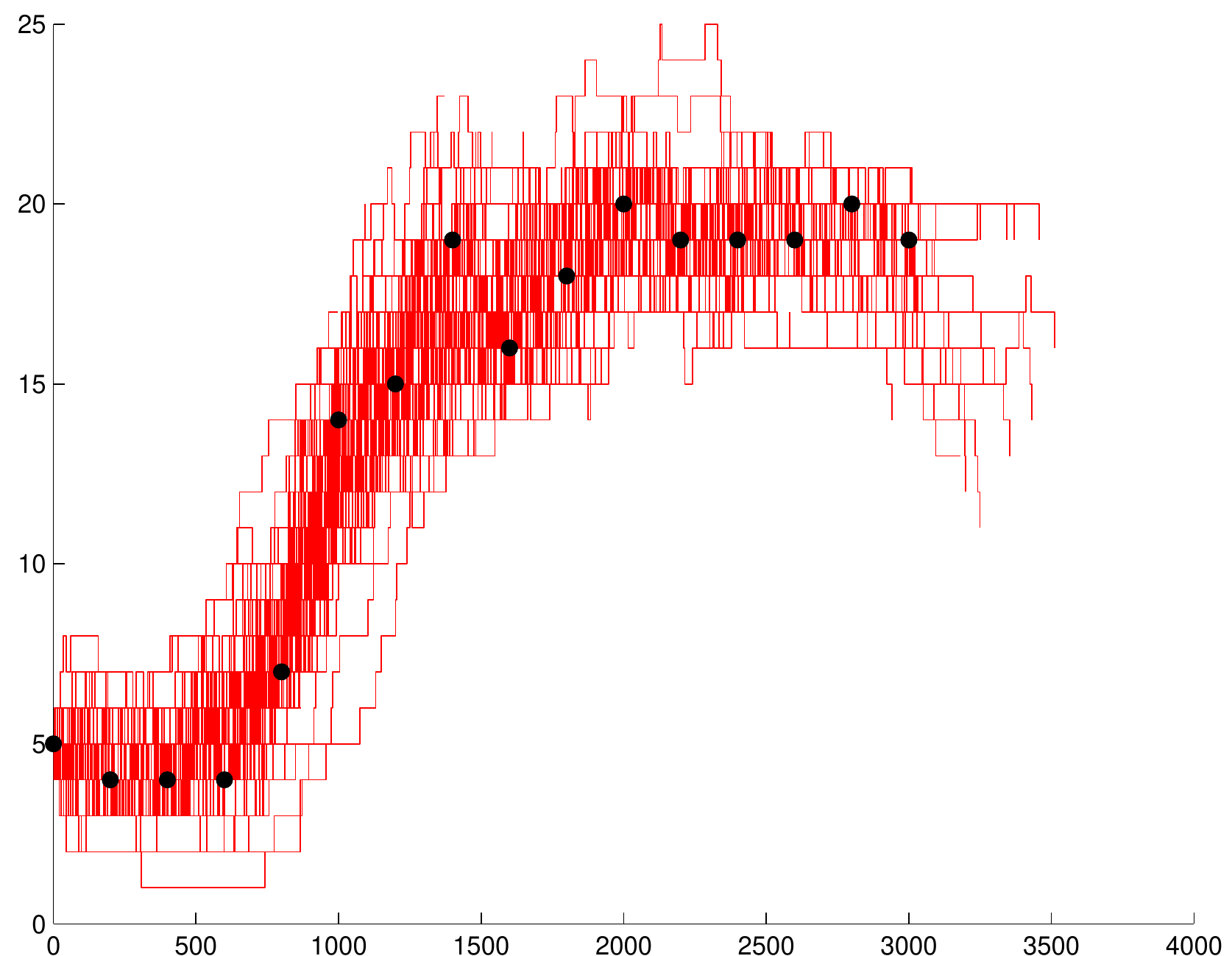}
		\end{subfigure}
		\caption{}
		\label{fig:traces}
	\end{subfigure}
	\caption{(a) Posterior marginals and pairwise correlations for the parameters of the LV model, from 5000 samples using Algorithm 2 (true values marked by red line, prior shown in dashed line); (b)~Samples of the posterior process: prey (top), predators (bottom). Dots indicate the observations.}
\end{figure*}

Results on the LV model show that methods based on random truncations achieve very considerable  improvements in performance compared to the Gibbs sampler (where the state space was truncated at a maximum number of 100 individuals per species). In particular, Algorithm 2 shows excellent behaviour in most aspects, with a high ESS suggesting it is a more efficient sampler. The running time of Algorithm 2 is comparable to that reported for a variational mean field approximation in \cite{opperVariational}, and its rapid convergence time suggests that this is a very competitive algorithm in practice.
Sample results from Algorithm 2 are presented in Figure~\ref{fig:panels} for the reaction parameters, and in Figure~\ref{fig:traces} for the state of the process itself. Algorithm 1, while still computationally feasible, requires a long time to converge, reflecting potential difficulties in choosing effective proposal distributions (a problem naturally bypassed by Algorithm 2).
The simple Gibbs algorithm is much slower than the other two, undoubtedly owing to its large state-space of 10000 states and very high memory requirements during the FFBS algorithm. Note that the impact of the (necessarily) large truncation is twofold. Firstly, the large state-space directly affects the running time of the FFBS algorithm, whose complexity is quadratic in the number of states. Secondly, since the rates in this model are increasing functions, having states with high counts means the generator matrix has high diagonal entries (exit rates). This, in turn, requires choosing a high exit rate for uniformisation, leading to long paths with many self-jumps, and ultimately further slowing down the FFBS step. The results for this model clearly show the usefulness of the random truncation approach compared to using a static, conservative truncation.


Results on the SIR model show that, in the finite state space case, the Gibbs sampler of Section 3.1 is highly efficient and by some way the best algorithm. This is unsurprising, as truncations incur additional computational overheads which are not needed for such a small state space.
The picture is completely different for the infinite SIR model.
In this case, the M-H sampler clearly seems to be the best algorithm, achieving very fast convergence and outperforming the other two. For parameter values within the prior range, the infinite SIR model exhibits fast dynamics which lead to very long uniformised trajectories, considerably increasing the computational costs of sampling trajectories via the FFBS algorithm. The problem is further compounded for the simple Gibbs sampler algorithm of Section~\ref{sec:inf}.
Even with the truncation described above, there are 32594 states, resulting in very severe computational and storage costs.

%


\subsection{Genetic toggle switch}
As a real application of our approach, we consider a model of a synthetic biological circuit describing the interaction between two genes (G1 and G2) and the proteins they encode (P1 and P2). Each protein acts as a repressor for the other gene, inhibiting its expression. This leads to a bistable behaviour, switching between a state with high P1 and low P2, and one with low P1 and high P2 (hence the name \emph{toggle-switch}). The interactions are encoded as eight chemical reactions:
\[
\begin{array}{lcll}
G_{1,\mathit{on}} &\rightarrow& G_{1,\mathit{on}} + P_1 & \text{at rate } \theta_1\\
G_{2,\mathit{on}} &\rightarrow& G_{2,\mathit{on}} + P_2 & \text{at rate } \theta_2\\
P_1 &\rightarrow& \varnothing & \text{at rate } \theta_3 P_1\\
P_2 &\rightarrow& \varnothing & \text{at rate } \theta_4 P_1\\
G_{1,\mathit{off}} &\rightarrow& G_{1,\mathit{on}} & \text{at rate } \theta_5\\
G_{2,\mathit{off}} &\rightarrow& G_{2,\mathit{on}} & \text{at rate } \theta_6\\
G_{1,\mathit{on}} &\rightarrow& G_{1,\mathit{off}} & \text{at rate } \theta_7 e^{rP_2}\\
G_{2,\mathit{on}} &\rightarrow& G_{2,\mathit{off}} & \text{at rate } \theta_8 e^{rP_1}\\
\end{array}
\]
where $r$ is a constant assumed known.

This system was engineered \textit{in vivo} in one of the pioneering studies in synthetic biology~\cite{gardner2000construction} and has been further studied in~\cite{Tian30052006}. Statistical inference is increasingly being recognised as a crucial bottleneck in synthetic biology: while genome engineering technologies enable researchers to reliably synthesise circuits with a desired structure, predicting the dynamic behaviour of a circuit requires knowledge of the kinetic parameters of the system once it is implanted {\it in the cell}, which cannot be directly measured. As synthetic biology is intrinsically at the single cell level, inference techniques for stochastic models have the potential to be of great aid in the rational design of synthetic biology circuits.


Following \cite{Tian30052006}, we model the system using a binary state for each gene and discrete levels for the proteins. The genes can be active or inactive, with protein being produced only in the former case. Each gene can be modelled with a telegraph process: an inactive gene becomes active at a constant rate, and an active one becomes inactive at a rate depending on the level of its repressor. When a gene is active, the level of its product follows a birth-death process; that is, proteins are produced at a constant rate and degrade at mass-action rates. We use a single production reaction for each protein to abstract various underlying mechanisms, including transcription and translation. The model comprises eight types of reaction; note that the requirements of our method on the form of the kinetic laws (Section~\ref{sec:inf}) are flexible enough to accommodate the deactivation dynamics used here, even though they are not mass-action.
\change{We simulated the system to produce behaviour similar to the simulated traces in~\cite{Tian30052006}.
	We kept 20 time points of measurements, which varied between 0 and 24 for each observed protein.}

\begin{figure*}
	\begin{subfigure}[b]{.4\linewidth}
		\includegraphics[scale=0.5]{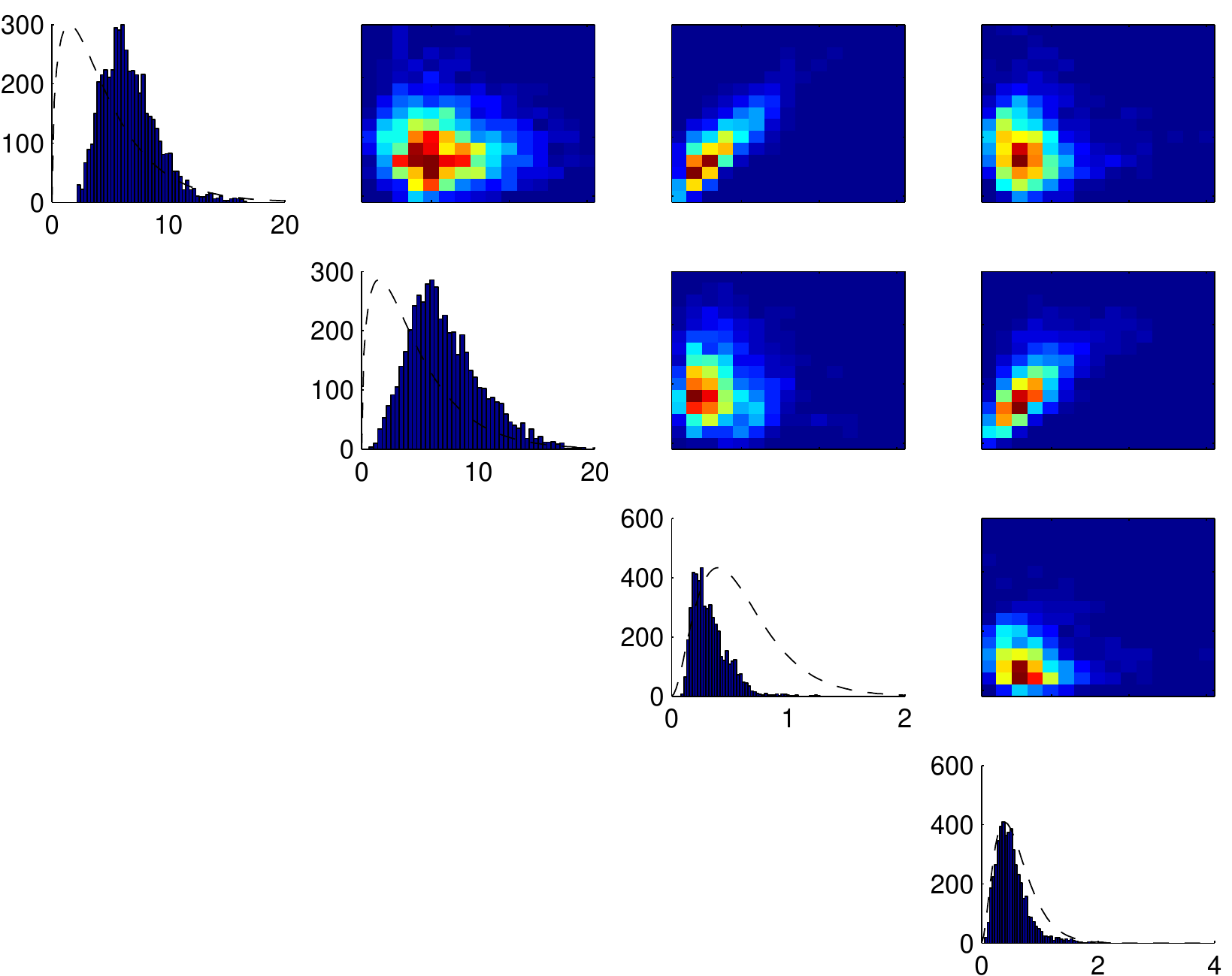} 
		\caption{}
	\end{subfigure}
	\hfill
	\begin{subfigure}[b]{.4\linewidth}
		\includegraphics[scale=0.38]{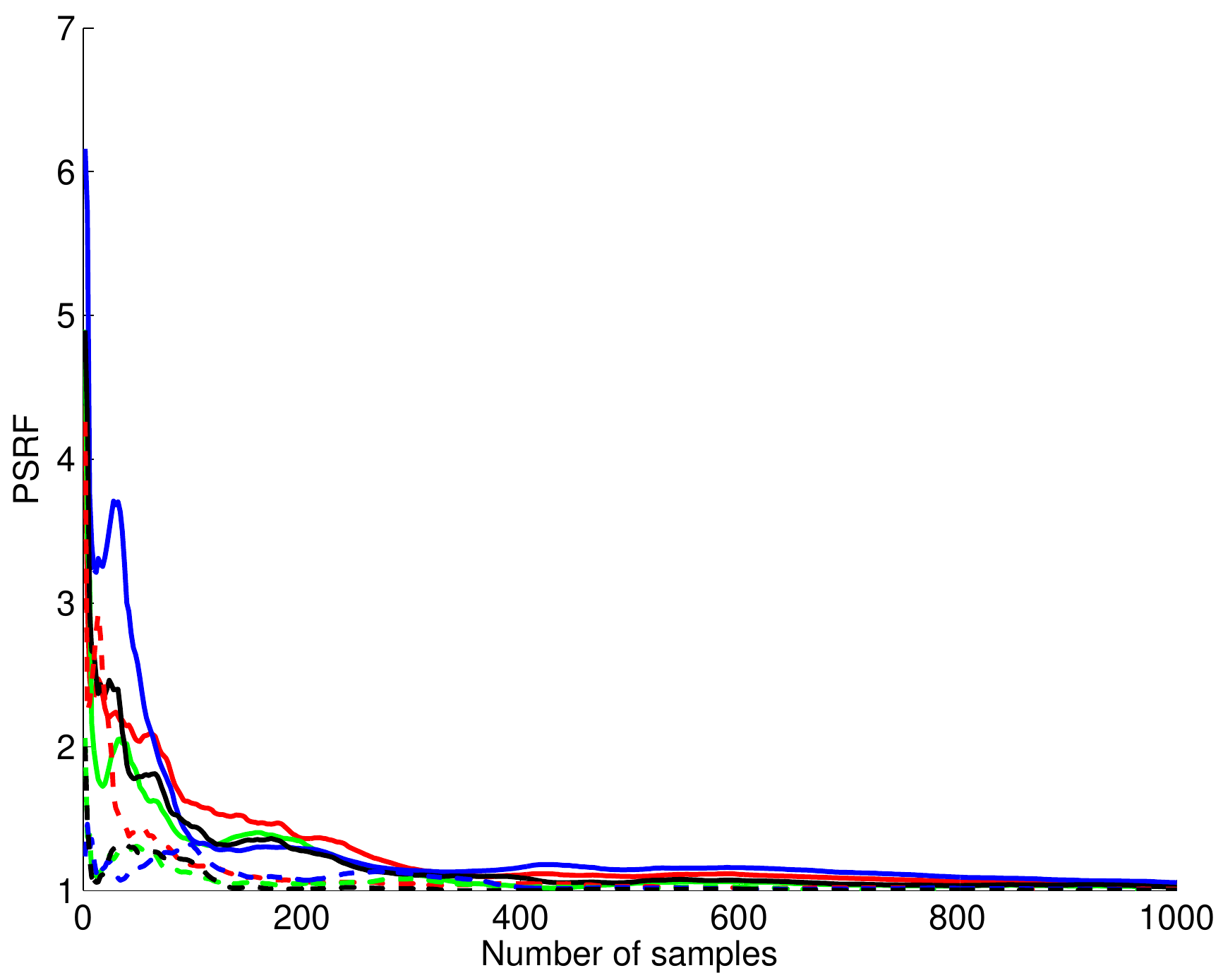}
		\caption{}
	\end{subfigure}
	\caption{(a) Posterior marginals and pairwise correlations for four parameters ($\theta_1$, $\theta_2$, $\theta_3$ and $\theta_4$) of the toggle switch model, from 5000 samples using Algorithm 2 (priors shown in dashed line); (b) PSRF for all eight parameters.}
	\label{fig:togglePlots}
\end{figure*}

We used Algorithm 2 to infer the joint posterior distribution of the eight parameters and state trajectories in this system. Our results indicate that the likelihood is relatively insensitive to the parameters governing the activation and deactivation of the two genes. This is a reasonable result, since we do not observe the state of the genes but only the levels of the two protein products. Therefore, the effect of the switching parameters is seen only indirectly through the switching events, which are rare in the data. In contrast, the protein expression and degradation rates have sharp posteriors which capture interesting correlations between the parameters --- for instance, we observe a strong correlation between the production and degradation rate of each protein, as perhaps expected given the similarity to a birth-death process. Figure~\ref{fig:togglePlots} shows parameter posteriors and convergence statistics for one such experiment, showcasing the good behaviour of the algorithm. 

\section{Related work}
\label{sec:related}
\change{Parameter inference in pMJPs has been the subject of previous work, with a significant body of literature focusing on continuous approximations to the process, in order to work around the complexities entailed by the stochastic dynamics.
	In general, such approximations are more accurate when the populations involved are high, and their accuracy degrades for lower populations as the impact of discrete stochastic behaviour becomes more pronounced. Two general classes of methods have been proposed to this end. The first involves approximating a pMJP with a diffusion process, as in~\cite{BIOM:BIOM345}, and using the resulting stochastic differential equations to calculate the likelihood. The second approach uses van Kampen's Linear Noise Approximation~\cite{vanKampen}, which assumes that the marginal distribution of the approximating process at any time is Gaussian. Under this assumption, ordinary differential equations for the mean and covariance can be derived as in~\cite{ruttor2009,BIOM:BIOM12152,komorowski2009} and used to compute the likelihood as part of an inference scheme.
	In contrast to these methods, our suggested approach is expected to be more accurate for smaller populations, as it maintains the stochastic dynamics. This makes it particularly useful for a range of systems which are large enough that a direct solution is inefficient, but not as large as to be accurately represented with continuous dynamics.}

In addition to MCMC-based approaches, like ours, particle methods have also been proposed for use with pMJPs, either with the exact dynamics~\cite{hajiaghayi2014,zechner2014scalable} or with continuous approximations such as the ones mentioned above.
However, they do require more user choices (e.g. number of particles) and can also incur heavy computational overheads for large models or state spaces. For infinite MJPs, in particular, the transition kernel is not available explicitly, making particle methods non-trivial and intrinsically expensive.
Variational methods have been developed in~\cite{opperVariational}, and can offer computational savings; however, the work in \cite{opperVariational} only performed state inference, providing point estimates for parameters. Furthermore, the error introduced by the variational approximation is often difficult to quantify.

\change{Recent work has made use of random truncations in different contexts:
	Strathmann \textit{et al.}~\cite{strathmann2015unbiased} propose using a Russian Roulette-style approach in large data scenarios where computing the likelihood from all data points is impractical, while Filippone \& Engler~\cite{FilipponeICML15} exploit the methodology to perform efficient inference for Gaussian processes.
	More generally, the construction of unbiased estimators has been the subject of theoretical and practical analysis.
	McLeish~\cite{mcleish2010general} and Rhee \& Glynn~\cite{rhee2013unbiased} examine the use of a method similar to Russian Roulette for obtaining unbiased estimates from biased ones.
	Agapiou \textit{et al.}~\cite{agapiou2014unbiased} consider ways of debiasing the estimates obtained by MCMC methods, particularly focusing on infinite spaces.
	Jacob \& Thiery~\cite{jacob2015} examine the theoretical existence of estimators that are both unbiased and guaranteed to be non-negative under different generation schemes.}

\section{Conclusions}
\label{sec:concl}
MJPs are common models in many branches of science, yet they still present fundamental statistical challenges. In this paper, we have proposed a novel MCMC framework for asymptotically exact inference for pMJPs, an important class of MJPs widely used in chemistry and systems biology. We remark that, while our focus is on biological applications, models with exactly the same structure are employed in many other fields, from epidemiology to ecology to performance modelling. Our random truncations pseudo-marginal approach enables a principled treatment of systems with potentially unbounded state-spaces. Interestingly, our results show that random truncations can also bring computational benefits over the naive alternative of bounding the state-space \textit{ab initio}, as done in \cite{RaoTehLong}. Intuitively, this is because choosing a truncation which guarantees a certain error bound usually requires still retaining a large state-space, while our random truncation method generally samples from much smaller systems.
The two truncation-based algorithms we consider here appear to perform best in different kinds of systems, and so neither can be said to be clearly superior in the general case.

\change{The performance of our proposed methods may vary with the system in question. As the number of species grows, the state-space grows exponentially larger, leading to increased computational overheads for our method (as for many other methods). While this may be a serious limitation for large models, it is worth pointing out that many practical applications of pMJPs describe systems with a small number of species, where our method's performance should not be affected. High counts of the species involved also result in larger state-spaces, leading to heavier computations, particularly for Algorithm 1. For Algorithm 2, the rates of the reactions can also have an impact: very fast reactions lead to a fine time-discretization and slower computations in the forward-backward step. Our methods perform best when particle numbers are not exceedingly large (otherwise, a continuous approximation would be both accurate and more efficient) and when observations are relatively dense or, equivalently, the process is not too volatile (or a truncation with many terms would be required for a good result).}

Pseudo-marginal methods based on random truncations are relatively new to statistics and machine learning~\cite{Rhee2012,girolami2013playing,FilipponeICML15}: to our knowledge, this is the first time that they are employed as a way of truncating an unbounded state space, and we think this idea may be appealing in other scenarios where unbounded state spaces are normal, such as non-parametric Bayesian methods. Compared to pseudo-marginal methods based on importance sampling~\cite{Filippone:pseudo14,hajiaghayi2014}, random truncations offer several advantages: there is no need to choose a proposal distribution, a notoriously difficult problem in high dimensions. \change{The choice of the truncating distribution, which controls the variance of the estimator, can in general be aided by some initial exploratory runs with different truncation distributions with different expected numbers of retained terms.} \change{Recent work on improving the behaviour of pseudo-marginal MCMC methods~\cite{murray2016aistats} may also be relevant to enhancing the performance of our proposed method.}


\paragraph*{Acknowledgements}
JH acknowledges support from the EU FET-Proactive programme through QUANTICOL grant 600708. GS acknowledges support from the European Research Council through grant MLCS306999. This work was supported by Microsoft Research through its PhD Scholarship Programme. The authors thank Maurizio Filippone, Mark Girolami and Chris Sherlock for useful discussions, Yichuan Zhang and Vinayak Rao for providing code and advice on computations, and the anonymous reviewers for their helpful feedback.

\bibliographystyle{unsrtnat}      
\bibliography{statcomp}

\begin{thebibliography}{34}
\providecommand{\natexlab}[1]{#1}
\providecommand{\url}[1]{\texttt{#1}}
\expandafter\ifx\csname urlstyle\endcsname\relax
  \providecommand{\doi}[1]{doi: #1}\else
  \providecommand{\doi}{doi: \begingroup \urlstyle{rm}\Url}\fi

\bibitem[Gardiner(1985)]{gardiner}
Crispin~W Gardiner.
\newblock \emph{Handbook of Stochastic Methods for Physics, Chemistry and the
  Natural Sciences, vol. 13 of}.
\newblock Springer Series in Synergetics, 1985.

\bibitem[Opper and Sanguinetti(2008)]{opperVariational}
Manfred Opper and Guido Sanguinetti.
\newblock {Variational inference for Markov jump processes}.
\newblock In J.C. Platt, D.~Koller, Y.~Singer, and S.~Roweis, editors,
  \emph{Advances in Neural Information Processing Systems 2008}, pages
  1105--1112. MIT Press, Cambridge, MA, 2008.

\bibitem[Cohn et~al.(2010)Cohn, El-Hay, Friedman, and Kupferman]{cohn2010mean}
Ido Cohn, Tal El-Hay, Nir Friedman, and Raz Kupferman.
\newblock {Mean Field Variational Approximation for Continuous-Time Bayesian
  Networks}.
\newblock \emph{The Journal of Machine Learning Research}, 11:\penalty0
  2745--2783, 2010.

\bibitem[Zechner et~al.(2014)Zechner, Unger, Pelet, Peter, and
  Koeppl]{zechner2014scalable}
Christoph Zechner, Michael Unger, Serge Pelet, Matthias Peter, and Heinz
  Koeppl.
\newblock Scalable inference of heterogeneous reaction kinetics from pooled
  single-cell recordings.
\newblock \emph{Nature Methods}, 11\penalty0 (2):\penalty0 197--202, 2014.

\bibitem[Hajiaghayi et~al.(2014)Hajiaghayi, Kirkpatrick, Wang, and
  Bouchard-C{\^o}t{\'e}]{hajiaghayi2014}
Monir Hajiaghayi, Bonnie Kirkpatrick, Liangliang Wang, and Alexandre
  Bouchard-C{\^o}t{\'e}.
\newblock {Efficient Continuous-Time Markov Chain Estimation}.
\newblock In \emph{Proceedings of the 31st International Conference on Machine
  Learning (ICML-14)}, pages 638--646, 2014.

\bibitem[Rao and Teh(2013)]{RaoTehLong}
Vinayak Rao and Yee~Whye Teh.
\newblock Fast {MCMC} sampling for {M}arkov jump processes and extensions.
\newblock \emph{Journal of Machine Learning Research}, 14:\penalty0 3207--3232,
  2013.
\newblock arXiv:1208.4818.

\bibitem[Lyne et~al.(2015)Lyne, Girolami, Atchadé, Strathmann, and
  Simpson]{girolami2013playing}
Anne-Marie Lyne, Mark Girolami, Yves Atchadé, Heiko Strathmann, and Daniel
  Simpson.
\newblock {On Russian Roulette Estimates for Bayesian Inference with
  Doubly-Intractable Likelihoods}.
\newblock \emph{Statist. Sci.}, 30\penalty0 (4):\penalty0 443--467, 11 2015.
\newblock \doi{10.1214/15-STS523}.
\newblock URL \url{http://dx.doi.org/10.1214/15-STS523}.

\bibitem[Filippone and Engler(2015)]{FilipponeICML15}
Maurizio Filippone and Raphael Engler.
\newblock Enabling scalable stochastic gradient-based inference for {G}aussian
  processes by employing the {U}nbiased {LI}near {S}ystem {S}olv{E}r
  ({ULISSE}).
\newblock In \emph{Proceedings of the 32nd International Conference on Machine
  Learning, ICML 2015, Lille, France, July 6-11, 2015}, 2015.

\bibitem[Al-Mohy and Higham(2011)]{exponentialAction}
Awad~H. Al-Mohy and Nicholas~J. Higham.
\newblock {Computing the Action of the Matrix Exponential, with an Application
  to Exponential Integrators}.
\newblock \emph{SIAM Journal on Scientific Computing}, 33\penalty0
  (2):\penalty0 488--511, 2011.
\newblock \doi{10.1137/100788860}.
\newblock URL \url{http://dx.doi.org/10.1137/100788860}.

\bibitem[Jensen(1953)]{jensen1953markoff}
Arne Jensen.
\newblock {Markoff chains as an aid in the study of Markoff processes}.
\newblock \emph{Scandinavian Actuarial Journal}, 1953\penalty0 (sup1):\penalty0
  87--91, 1953.

\bibitem[Munsky and Khammash(2006)]{fsp}
Brian Munsky and Mustafa Khammash.
\newblock The finite state projection algorithm for the solution of the
  chemical master equation.
\newblock \emph{The Journal of Chemical Physics}, 124\penalty0 (4):\penalty0
  044104, 2006.
\newblock \doi{http://dx.doi.org/10.1063/1.2145882}.
\newblock URL
  \url{http://scitation.aip.org/content/aip/journal/jcp/124/4/10.1063/1.2145882}.

\bibitem[Andrieu and Roberts(2009)]{andrieu2009pseudo}
Christophe Andrieu and Gareth~O. Roberts.
\newblock {The pseudo-marginal approach for efficient Monte Carlo
  computations}.
\newblock \emph{The Annals of Statistics}, pages 697--725, 2009.

\bibitem[Beaumont(2003)]{Beaumont1139}
Mark~A. Beaumont.
\newblock Estimation of population growth or decline in genetically monitored
  populations.
\newblock \emph{Genetics}, 164\penalty0 (3):\penalty0 1139--1160, 2003.
\newblock ISSN 0016-6731.
\newblock URL \url{http://www.genetics.org/content/164/3/1139}.

\bibitem[Doucet et~al.(2015)Doucet, Pitt, Deligiannidis, and
  Kohn]{Doucet07032015}
A.~Doucet, M.~K. Pitt, G.~Deligiannidis, and R.~Kohn.
\newblock {Efficient implementation of Markov chain Monte Carlo when using an
  unbiased likelihood estimator}.
\newblock \emph{Biometrika}, 2015.
\newblock \doi{10.1093/biomet/asu075}.
\newblock URL
  \url{http://biomet.oxfordjournals.org/content/early/2015/03/07/biomet.asu075.abstract}.

\bibitem[Sherlock et~al.(2015)Sherlock, Thiery, Roberts, and
  Rosenthal]{sherlock2015}
Chris Sherlock, Alexandre~H. Thiery, Gareth~O. Roberts, and Jeffrey~S.
  Rosenthal.
\newblock {On the efficiency of pseudo-marginal random walk Metropolis
  algorithms}.
\newblock \emph{Ann. Statist.}, 43\penalty0 (1):\penalty0 238--275, 02 2015.
\newblock \doi{10.1214/14-AOS1278}.
\newblock URL \url{http://dx.doi.org/10.1214/14-AOS1278}.

\bibitem[Rhee and Glynn()]{rhee2013unbiased}
CH~Rhee and Peter~W Glynn.
\newblock Unbiased estimation with square root convergence for sde models.
\newblock \emph{To appear}.
\newblock URL \url{http://rhee.gatech.edu/papers/RheeGlynn13a.pdf}.

\bibitem[Kleinrock(1975)]{kleinrock}
Leonard Kleinrock.
\newblock \emph{{Queueing Systems}}, volume I: Theory.
\newblock Wiley Interscience, 1975.

\bibitem[Boys et~al.(2008)Boys, Wilkinson, and Kirkwood]{boys2008bayesian}
Richard~J Boys, Darren~J Wilkinson, and Thomas~BL Kirkwood.
\newblock Bayesian inference for a discretely observed stochastic kinetic
  model.
\newblock \emph{Statistics and Computing}, 18\penalty0 (2):\penalty0 125--135,
  2008.

\bibitem[Anderson and May(1991)]{anderson1991infectious}
Roy~M Anderson and Robert~McCredie May.
\newblock \emph{Infectious diseases of humans}, volume~1.
\newblock Oxford university press Oxford, 1991.

\bibitem[Gelman et~al.(2014)Gelman, Carlin, Stern, and
  Rubin]{gelman2014bayesian}
Andrew Gelman, John~B Carlin, Hal~S Stern, and Donald~B Rubin.
\newblock \emph{Bayesian Data Analysis}.
\newblock Taylor \& Francis, 2014.

\bibitem[Gardner et~al.(2000)Gardner, Cantor, and
  Collins]{gardner2000construction}
Timothy~S Gardner, Charles~R Cantor, and James~J Collins.
\newblock {Construction of a genetic toggle switch in Escherichia coli}.
\newblock \emph{Nature}, 403\penalty0 (6767):\penalty0 339--342, 2000.

\bibitem[Tian and Burrage(2006)]{Tian30052006}
Tianhai Tian and Kevin Burrage.
\newblock Stochastic models for regulatory networks of the genetic toggle
  switch.
\newblock \emph{Proceedings of the National Academy of Sciences}, 103\penalty0
  (22):\penalty0 8372--8377, 2006.
\newblock \doi{10.1073/pnas.0507818103}.

\bibitem[Golightly and Wilkinson(2005)]{BIOM:BIOM345}
A.~Golightly and D.~J. Wilkinson.
\newblock {Bayesian Inference for Stochastic Kinetic Models Using a Diffusion
  Approximation}.
\newblock \emph{Biometrics}, 61\penalty0 (3):\penalty0 781--788, 2005.
\newblock ISSN 1541-0420.
\newblock \doi{10.1111/j.1541-0420.2005.00345.x}.
\newblock URL \url{http://dx.doi.org/10.1111/j.1541-0420.2005.00345.x}.

\bibitem[Van~Kampen(1992)]{vanKampen}
Nicolaas~Godfried Van~Kampen.
\newblock \emph{Stochastic processes in physics and chemistry}, volume~1.
\newblock Elsevier, 1992.

\bibitem[Ruttor and Opper(2009)]{ruttor2009}
Andreas Ruttor and Manfred Opper.
\newblock {Efficient Statistical Inference for Stochastic Reaction Processes}.
\newblock \emph{Physical review letters}, 103\penalty0 (23), 2009.
\newblock \doi{10.1103/PhysRevLett.103.230601}.
\newblock URL \url{http://link.aps.org/doi/10.1103/PhysRevLett.103.230601}.

\bibitem[Fearnhead et~al.(2014)Fearnhead, Giagos, and Sherlock]{BIOM:BIOM12152}
Paul Fearnhead, Vasilieos Giagos, and Chris Sherlock.
\newblock Inference for reaction networks using the linear noise approximation.
\newblock \emph{Biometrics}, 70\penalty0 (2):\penalty0 457--466, 2014.
\newblock ISSN 1541-0420.
\newblock \doi{10.1111/biom.12152}.
\newblock URL \url{http://dx.doi.org/10.1111/biom.12152}.

\bibitem[Komorowski et~al.(2009)Komorowski, Finkenst{\"a}dt, Harper, and
  Rand]{komorowski2009}
Micha{\l} Komorowski, B{\"a}rbel Finkenst{\"a}dt, Claire Harper, and David
  Rand.
\newblock Bayesian inference of biochemical kinetic parameters using the linear
  noise approximation.
\newblock \emph{BMC bioinformatics}, 10\penalty0 (1):\penalty0 343, 2009.
\newblock \doi{10.1186/1471-2105-10-343}.

\bibitem[Strathmann et~al.(2015)Strathmann, Sejdinovic, and
  Girolami]{strathmann2015unbiased}
Heiko Strathmann, Dino Sejdinovic, and Mark Girolami.
\newblock Unbiased bayes for big data: Paths of partial posteriors.
\newblock \emph{arXiv preprint arXiv:1501.03326}, 2015.

\bibitem[McLeish(2011)]{mcleish2010general}
Don McLeish.
\newblock A general method for debiasing a monte carlo estimator.
\newblock \emph{Monte Carlo Methods and Applications}, 2011.

\bibitem[Agapiou et~al.(2014)Agapiou, Roberts, and
  Vollmer]{agapiou2014unbiased}
Sergios Agapiou, Gareth~O Roberts, and Sebastian~J Vollmer.
\newblock Unbiased monte carlo: posterior estimation for
  intractable/infinite-dimensional models.
\newblock \emph{arXiv preprint arXiv:1411.7713}, 2014.

\bibitem[Jacob and Thiery(2015)]{jacob2015}
Pierre~E. Jacob and Alexandre~H. Thiery.
\newblock On nonnegative unbiased estimators.
\newblock \emph{Ann. Statist.}, 43\penalty0 (2):\penalty0 769--784, 04 2015.
\newblock \doi{10.1214/15-AOS1311}.
\newblock URL \url{http://dx.doi.org/10.1214/15-AOS1311}.

\bibitem[Rhee and Glynn(2012)]{Rhee2012}
Chang-han Rhee and Peter~W. Glynn.
\newblock A new approach to unbiased estimation for sde's.
\newblock In \emph{Proceedings of the Winter Simulation Conference}, WSC '12,
  pages 17:1--17:7, 2012.
\newblock URL \url{http://dl.acm.org/citation.cfm?id=2429759.2429780}.

\bibitem[Filippone and Girolami(2014)]{Filippone:pseudo14}
Maurizio Filippone and Mark~A. Girolami.
\newblock {Pseudo-Marginal Bayesian Inference for Gaussian Processes}.
\newblock \emph{{IEEE} Trans. Pattern Anal. Mach. Intell.}, 36\penalty0
  (11):\penalty0 2214--2226, 2014.
\newblock \doi{10.1109/TPAMI.2014.2316530}.
\newblock URL
  \url{http://doi.ieeecomputersociety.org/10.1109/TPAMI.2014.2316530}.

\bibitem[Murray and Graham()]{murray2016aistats}
Iain Murray and Matthew~M. Graham.
\newblock Pseudo-marginal slice sampling.
\newblock \emph{AISTATS 2016 (to appear), arXiv preprint arXiv:1510.02958}.

\end{thebibliography}

\end{document}